# کنترل ترافیک با استفاده از زمانبندی هوشمند چراغ های راهنمایی رانندگی با تکنیک یادگیری تقویتی و پردازش بلادرنگ تصاویر دوربین های نظارتی

مهدی جامه بزرگ[1]، محسن حامی[2,*]، سجاد ده ده جانی[3]

1- دانشجوی کارشناسی مهندسی کامپیوتر دانشگاه بو علی سینا

2- فارغ التحصیل کارشناسی مهندسی کامپیوتر دانشگاه بو علی سینا

3- فارغ التحصیل ارشد حقوق جزا و جرم شناسی دانشگاه آزاد واحد بوییین زهرا،مدرس دانشگاه امین

**چکیده**

زمانبندی بهینه چراغ های راهنمایی و رانندگی از عوامل بسیار اثر گذار بر کاهش ترافیک شهری است، در اغلب سیستم های قدیمی از زمانبندی ثابت برای کنترل ترافیک استفاده می شد که این موضوع از نظر زمان و هزینه کارایی چندانی ندارد، روش های امروزی در حوزه مدیریت ترافیک مبتنی بر استفاده از هوش مصنوعی است در این روش با استفاده از پردازش بلادرنگ تصاویر دوربین های نظارت تصویری به همراه یادگیری تقویتی، نتیجتا زمان بندی بهینه چراغ های راهنمایی و رانندگی بر اساس پارامتر های متعددی تعیین و اعمال می شود،در پژوهش از روش های یادگیری عمیق در تشخیص وسایل نقلیه از مدل YOLOv9-C استفاده شده تا تخمینی از تعداد و دیگر ویژگی های وسایل نقلیه مانند سرعت داشته باشیم سپس با مدلسازی وسایل نقلیه در شبیه ساز محیط شهری در OpenAI gym از یادگیری تقویتی چند عامله و الگوریتم DQN Rainbow به منظور زمانبندی چراغ های ترافیکی تقاطع ها استفاده شده است،همچنین آموزش دوباره مدل بر روی تصاویر ماشین های ایرانی باعث افزایش دقت شده است، نتایج روش پیشنهادی، مدلی است که در تحلیل دوربین های نظارتی و یافتن زمانبندی بهینه به دقت های بطور قابل ملاحظه بهتری نسبت به پژوهش های قبلی رسیده است.

**کلید واژه:** *مدیریت هوشمند ترافیک ،یادگیری تقویتی، پردازش تصویر،یادگیری عمیق، زمانبندی بهینه چراغ های ترافیکی*

---

1 محقق حوزه هوش مصنوعی، 989928618623+، mahdijamebozorg2000@gmail.com

2 محقق حوزه هوش مصنوعی، 989190981446+ ، m.hami@eng.basu.ac.ir (مسئول مکاتبات)

2 مدرس دانشگاه امین ، sajjad.dehgaani@gmail.com

## 1- مقدمه

ترافیک شهری از جمله معضلات زندگی روزمره افراد، بویژه در کلان شهر ها است راهکار حل این مشکل میتواند یک پژوهش بین رشته ای باشد که تمام جنبه های کنترل ترافیک را در نظر بگیرید، در کشور های توسعه یافته افراد زمان خود را بر اساس مدت زمانی که در ترافیک معطل می شوند تنظیم می کنند [1] ، از اثرات اولیه و محسوس ترافیک، معطلی افراد است اما این پدیده میتواند جنبه های زیست محیطی، بهداشتی و اقتصادی نیز داشته باشد که نامحسوس اما بسیار مهم هستند [2]،[3] امروزه با پیشرفت سیستم های پردازشی، استفاده از هوش مصنوعی در پیش بینی و کنترل ترافیک میتواند بسیار موثر عمل کند.

بررسی کلی انواع ازدحام و انواع دلایل رخداد ترافیک، یک دید کلی نسبت به مشکل و راه حل های موجود خواهد داد.

ازدحام پشت تقاطع ها و تشکیل صف های طولانی خودرو ها از پدیده های رایج امروز شهر هاست[4]، این ازدحام میتواند به دو صورت روزمره و غیر مترقبه رخ دهد. [5]

ازدحام روزمره ناشی از بیشتر بودن تقاضا نسبت به ظرفیت در ساعات اوج عبور و مرور است [6] در حالی که ازدحام غیر مترقبه ناشی از اتفاقات ناگهانی موثر بر ترافیک مانند تصادف،وضعیت آب و هوایی،نقص فنی، وضعیت بحرانی (مانند زلزله) و موارد اینچنین باشد [7]، هر چند احداث راه های جدید ، تعریض مسیر ها و افزایش ظرفیت معیار می تواند باعث کاهش ترافیک شهری شود[8] اما این قبیل از راه حل ها همراه با صرف هزینه های بسیار زیاد است. همچنین تجربه نشان داده است افزایش ظرفیت در نقطه ای از شهر میتواند باعث ایجاد ازدحام در مکان دیگری از شبکه ترافیکی شهر شود.[9]

راهکار پیشنهادی استفاده از سیستم های حمل نقل هوشمند است که از فناوری های مخابراتی،کنترلی و کامپیوتری برای مدیریت و کنترل موثر ترافیک بهر می برند ، این سیستم ها انعطاف پذیری را برای کنترل موثر ترافیک فراهم میکنند و عملکرد شبکه ترافیکی را با کنترل جریان ترافیکی بهبود می دهند.[10][11]

بطور کلی کنترل ترافیک موثر میتواند شامل مولفه متفاوتی باشد که شامل کنترل هوشمند چراغ های راهنمایی در تقاطع ها، کنترل جریان ورودی به بزرگراه ها،اعمال محدودیت سرعت در معابر مختلف،تابلو های متغییر خبری است. [12].

مکانیزم مدیریت ترافیک مبتنی بر کنترل چراغ های ترافیکی به دو دسته سیگنال مجزا و مرتبط تقسیم می شود.[13]

**روش سیگنال مجزا**: در این روش هر تقاطع بصورت جداگانه (بدون توجه به وضعیت سایر تقاطع ها) سیگنال ترافیک در نظر گرفته می شود

روش کنترل زمان ثابت : در این روش از کنترل کننده برای چراغ های هر تقاطع زمانی از قبل تعیین میکند و زمان چراغ ها بر اساس داده های ترافیکی و الگوی ترافیک تقاطع محاسبه می شود که این زمانبندی میتواند در ساعات مختلف روز متفاوت باشد.

**روش کنترل واکنشی** : در این روش تقاطع به سنسور هایی مجهز می شود تا وضعیت ترافیکی را بصورت بلادرنگ به کنترل کننده ارسال کند و کنترل کننده بر اساس اطلاعات دریافتی از سنسور ها زمان مناسب برای آن تقاطع را حساب می کند.

**روش سیگنال مرتبط** :در این روش یک دید کلی نسبت به معابر وجود داشته و اثر تقاطع های مختلف بر روی هم محاسبه شده و زمانبندی انجام می شود، بر همین اساس این روش با دو رویکرد کلی متمرکز و غیر متمرکز انجام می شود.

**رویکرد متمرکز(centralized)** : در این رویکرد داده ها تنها به یک سرور مرکزی ارسال می شود و این سرور پس از تصمیم گیری نتایج را به کنترل کننده های تقاطع ها ارسال می کند در این رویکرد بدلیل وجود تمام داده های ترافیکی در لحظه میتوان بهینه سازی های گسترده تری را انجام داده اما متمرکز بودن میتواند باعث ایجاد نقطه تکی خرابی ( single point of failure) شود یعنی با خرابی این سرور مرکزی تمام زمان بندی ها بی اثر خواهند شد.

۳

**رویکرد غیر متمرکز (decentralized):** فرضیه این روش بر این اساس است که تقاطع های دور از هم بر روی یکدیگر تاثیر ناچیزی می گذارند پس با توجه به گستردگی شبکه کنترل ترافیک میتوان معابر را ناحیه بندی کرد [14] ، که این موضوع منجر به توزیع بار بین کامپیوتر های محلی و افزایش سرعت پاسخگویی سرور ها هم می شود. هر چند بعضی از الگوریتم های یادگیری تقویتی برای حالت متمرکز طراحی شده اند اما در حالت تعدد تقاطع ها با توجه به زیاد بودن تعداد حالت ها تاخیر تصمیم گیری زیاد شده و فضای حالت های مدل بصورت نمایی افزایش می یابد که استفاده از اینترنت اشیا به همراه مدل های یادگیری ماشین توزیع شده باعث حل این مشکل شده است.[15]و[19] در این پژوهش این رویکرد مد نظر است.

## 2-روش تحقیق :

پیشرفت زمینه های مختلف تکنولوژی باعث بکار گیری سامانه های تمام هوشمند و خودکار در عمل شده است افزایش توان پردازشی سیستم های نهفته در حوزه اینترنت اشیا و از طرف دیگر سبکتر و دقیق تر شدن مدل های هوش مصنوعی در سال های اخیر امکان استفاده عملی را فراهم کرده است در این پژوهش ما با استفاده جدید ترین روش های و معماری های هوش مصنوعی سعی در طراحی یک سامانه یکپارچه برای مدیریت و نظارت ترافیک داریم.

**پردازش تصویر:** مدل یادگیری عمیق استفاده شده در پژوهش YOLOv9-C بوده که در تاریخ 2 اسفند 1402 منتشر شده است این مدل بهترین ابزار object detection در حوزه پردازش تصویر است ورژن با تعداد پارامتر های کمتر آن که تحت عنوان Tiny YOLO معروف است بدلیل سبک بودن از سرعت بالایی در استنتاج به ویژه در وظایف بلادرنگ برخوردار است. بطور مثال پژوهش [16] در سال 2019 در عمل به سرعت پردازشی 33.5 فریم بر ثانیه در کاربرد شمارش ماشین ها در تصاویر دوربین های نظارتی ترافیکی رسیده است که در مدل استفاده شده این پژوهش به عدد 55 فریم بر ثانیه رسیده است همچنین تعداد پارامتر های مدل 25.5 میلیون و میانگین صحت 53٪ می باشد که نسبت به نسل های قبل تر خود بسیار سبکتر و دقیق تر شده است.[17] [شکل3]

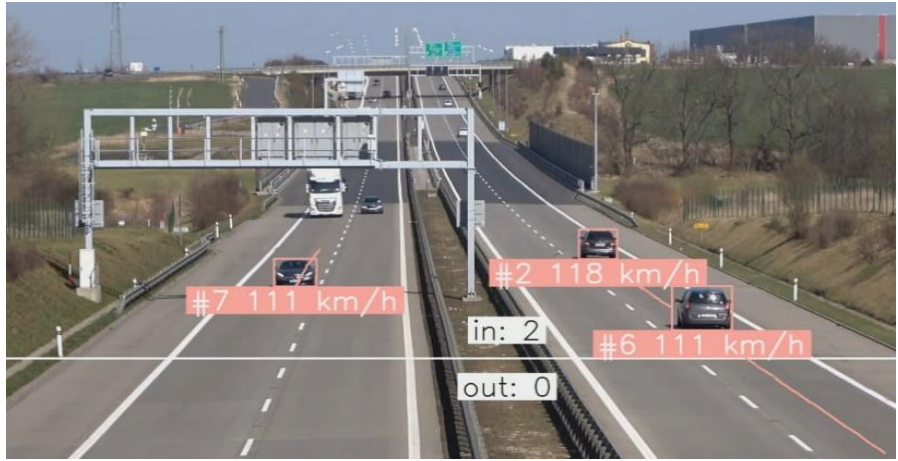

شکل 1: خروجی مدل YOLOv9 (وسایل نقلیه شمارش شده و شناسه یکتا و سرعت گرفته اند)

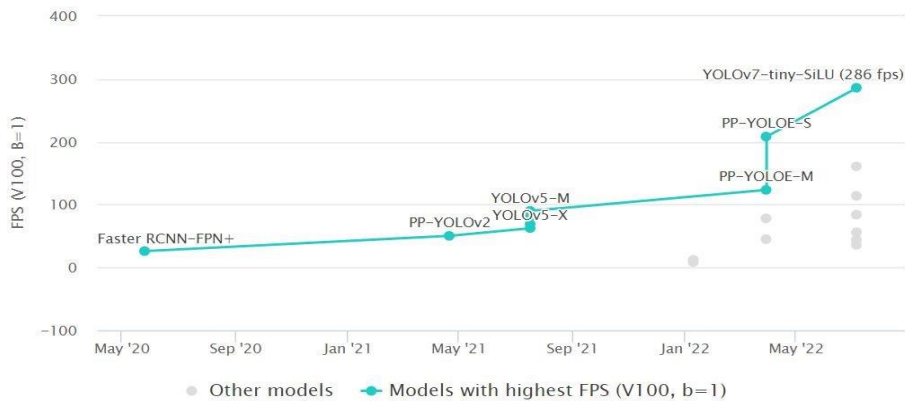

شکل 2: مقایسه تعداد فریم بر ثانیه پردازشی مدل های هوش مصنوعی تشخیص اشیا

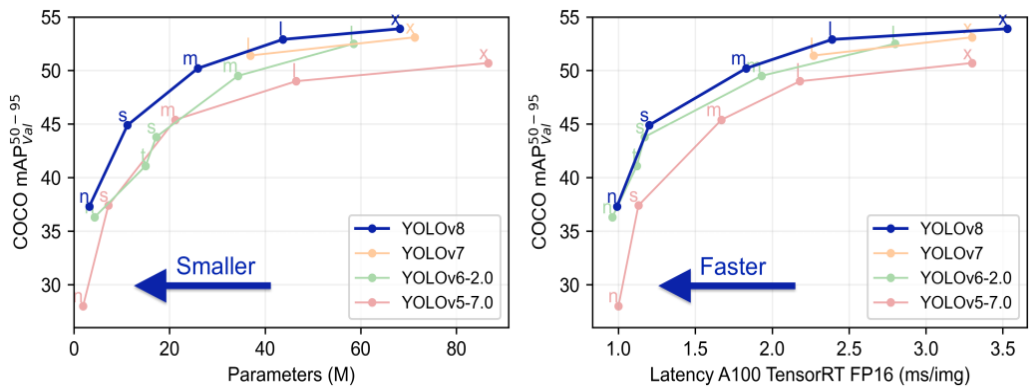

شکل 3: مقایسه تاخیر و تعداد پارامتر های مدل یادگیری عمیق سری YOLO

منبع شکل 2و3 : www.viso.ai , Object Detection in 2024: The Definitive Guide

۵

**هسته یادگیری تقویتی** : استفاده از الگوریتم مدرن Rainbow [18] در این مقاله می‌تواند به عنوان یک ابزار قدرتمند برای بهبود عملکرد و کارایی مدل‌های یادگیری تقویتی در مدیریت ترافیک موثر باشد. مزایای استفاده از این الگوریتم عبارتند از:

**۱. ترکیبی از چند روش:** الگوریتم Rainbow یک ترکیب موثر از چندین تکنیک یادگیری تقویتی مانند Double Q-Learning، Dueling Network، Prioritized Experience Replay و Distributional RL استفاده می‌کند. این ترکیب تنوع بیشتری در روش‌های آموزش و بهبود عملکرد مدل‌ها فراهم می‌کند.

**۲. کاهش انحراف واریانس:** از جمله مزیت‌های الگوریتم Rainbow، کاهش انحراف واریانس در تخمین ارزش عملکردها می‌باشد که می‌تواند بهبود قابل توجهی در پایداری و سرعت آموزش مدل‌های یادگیری تقویتی ایجاد کند.

**۳. انعطاف‌پذیری:** الگوریتم Rainbow از رویکردهای مختلفی برای بهبود عملکرد یادگیری تقویتی استفاده می‌کند و این انعطاف‌پذیری آن را به یک ابزار قدرتمند در مدیریت ترافیک تبدیل می‌کند که می‌تواند با توجه به شرایط و محیط‌های مختلف، به صورت بهینه تطبیق یابد.

$$reward = \sum avg(speeds)$$
$$+ \textit{The number of cars passing through the intersection}$$
$$- \sum \textit{fairness , remaining vechicle , stuck vechilce}$$

فرمول شماره 1: فرمول پیشنهادی محاسبه پاداش در الگوریتم یادگیری تقویتی

**Remaining vehicle:** تعداد ماشین های هر خیابون که طی چراغ سبز قبلی رد نشدن

**Stuck vehicle:** تعداد ماشین هایی که داخل تقاطع گیر کرده اند

**Fairness:** اختلاف بیشترین و کمتری میانگین زمان انتظار در هر کدام از راه ها

۶

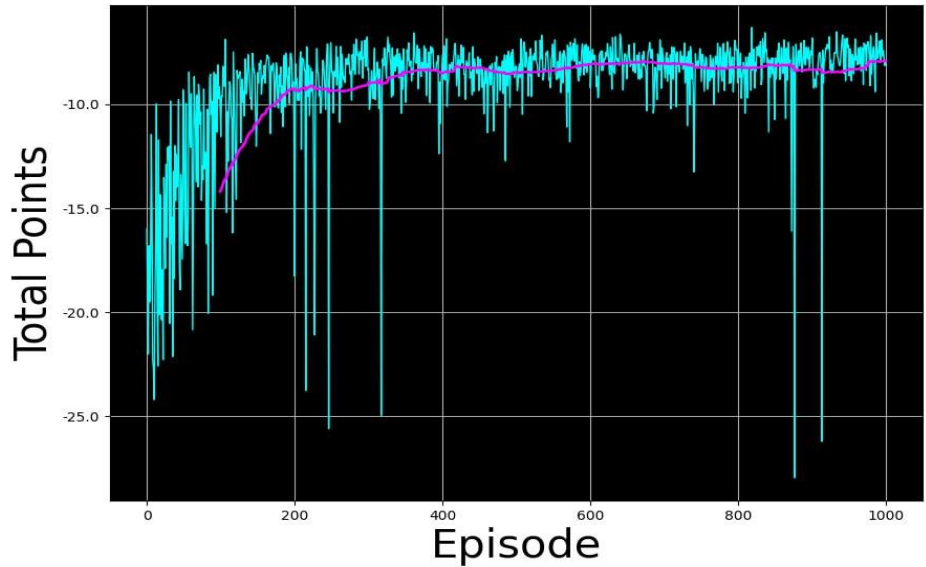

شکل 4: خروجی مدل از روند یادگیری الگوریتم DQN Rainbow در طول 1000 اپیزود

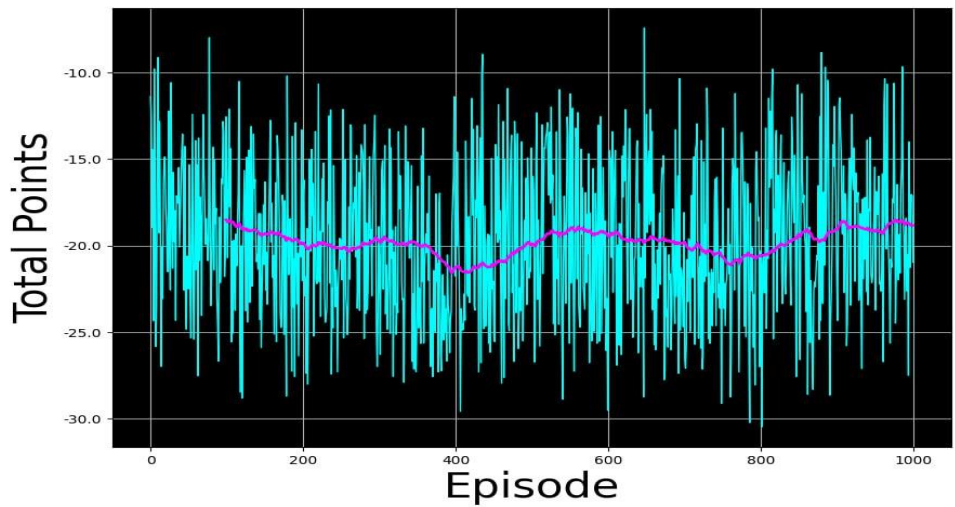

شکل 5: خروجی مدل از روند یادگیری الگوریتم DQN در طول 1000 اپیزود

مطابق شکل 3و4 الگوریتم DQN بصورت کلی یادگیری درستی نداشته و نتوانسته پاداش(reward) را بیشینه کند، اما الگوریتم DQN Rainbow توانسته بخوبی پاداش رو بشینه کند.

```
reward = 0
#Penalties
reward -= self.avg_waiting_times.sum()//(self.roads_count-1)
reward -=   remaining_vehicles.sum ()
#Fairness (wighted)
reward -= (np.max(self.avg_waiting_times) - 
np.min(self.avg_waiting_times))*2
#Reward
reward += self.in_counts.sum()
reward += self.out_counts.sum()
```

شبه کد شماره 1 : فرمول محاسبه پاداش و جریمه در الگوریتم پیشنهادی

**محیط شبیه ساز :** این محیط شبیه‌سازی یک شبیه‌سازی ساده از یک محیط مدیریت ترافیک است که برای استفاده در آموزش مدل‌های یادگیری تقویتی طراحی شده است. در این محیط، تعدادی خیابان و چراغ راهنمایی و رانندگی وجود دارد. برخی از ویژگی‌های این محیط عبارتند از:

**تعداد خیابان‌ها:** تعداد خیابان‌های موجود در محیط توسط پارامتر roads_count تعیین می‌شود.

**تغییرات در زمان:** با توجه به انتخاب اقدامات موجود در فضای عمل، زمان سبز و قرمز چراغ‌های راهنمایی تغییر می‌کند.

**فضای عمل:** فضای عمل این محیط توسط یک فضای عمل گسسته تعیین می‌شود که هر عمل یک انتخاب برای هر خیابان و تغییر کلی همزمان چراغ‌های راهنمایی را مشخص می‌کند.

**فضای مشاهدات:** این محیط دارای یک فضای مشاهدات است که اطلاعاتی از وضعیت فعلی محیط را ارائه می‌دهد، از جمله زمان باقی‌مانده چراغ‌های راهنمایی، متوسط زمان انتظار، تعداد وسایل نقلیه، تعداد وسایل نقلیه ورودی و خروجی و ...

**شبیه‌سازی:** با انجام هر اقدام، وضعیت جدید محیط به ازای اقدام انجام شده محاسبه شده و مشاهده می‌شود.

**پاداش:** سیستم پاداش در این محیط شامل مجموعه‌ای از تاثیرات مثبت و منفی بر روی وضعیت ترافیک و همچنین عدم تعادل در زمان انتظار برای هر خیابان است.

### 3- مروری بر کار های پیشین

در این بخش به بررسی مقاله ای متمایز چند سال اخیر می پردازیم ، هر چند بیشتر این مقالات کاربرد شبیه سازی دارند به این منظور که ترافیک غیر واقعی را بر اساس توزیع های آماری (مانند توزیع گاوسی) تولید می کنند که چندان دقیق نیست در پژوهش ما ترافیک واقعی بصورت بلادرنگ از طریق دربین های نظارت تصویری استخراج می شود.

در مقاله [20] یک چارچوب چندعاملی برای کنترل سیگنال‌های ترافیکی با استفاده از یادگیری تقویتی عمیق می‌دهد. نوآوری اصلی این مقاله در استفاده از شبکه‌های عصبی عمیق برای تصمیم‌گیری متمرکز بر کنترل ترافیک در محیط‌های پیچیده است.

در مقاله [21] به بررسی کنترل سطح انسانی از طریق یادگیری تقویتی عمیق می‌پردازد. نتایج این مقاله نشان می‌دهد که الگوریتم‌های یادگیری تقویتی عمیق می‌توانند به عنوان روش‌های موثر برای کنترل سیستم‌های پیچیده مورد استفاده قرار بگیرند.

### 4- نتیجه‌گیری

در این مقاله، یک روش نوین برای مدیریت ترافیک با استفاده از یادگیری تقویتی و پردازش تصویر با استفاده از آخرین دستاورد های این حوزه معرفی شد که با استفاده از داده‌های واقعی حاصل از دوربین‌های مدار بسته و مکانیزم تایمر، بهبودهای قابل توجهی در مدیریت ترافیک به دست آورده است. این روش نه‌تنها باعث افزایش بهره‌وری در تصمیم‌گیری‌ها و کاهش تغییرات ناگهانی در ترافیک می‌شود، بلکه همچنین با توجه به واقعیت‌های موجود در محیط ترافیک، به دقت بیشتری در مدیریت ترافیک دست پیدا می‌کند.

استفاده از این روش می‌تواند به بهبود عملکرد سیستم‌های ترافیکی شهری و کاهش زمان‌های ترافیکی و احتمال وقوع حوادث کمک کند. امیدواریم که این روش مورد استفاده قرار گیرد و نتایج مثبتی را در بهبود شرایط ترافیکی به ارمغان آورد.

# Traffic control using intelligent timing of traffic lights with reinforcement learning technique and real-time processing of surveillance camera images

**Mahdi Jamebozorg[1], Mohsen Hami[2],* and Sajjad Deh Deh Jani [3]**

1- undergraduate students in Computer Engineering at Bu-Ali Sina University, Iran

2- Bachelor's degree holder in Computer Engineering from Bu-Ali Sina University, Iran

3- master's degree holder in criminal law from Buen-Zahara Islamic Azad University, Iran

**Abstract**

Optimal management of traffic light timing is one of the most effective factors in reducing urban traffic. In most old systems, fixed timing was used along with human factors to control traffic, which is not very efficient in terms of time and cost. Nowadays, methods in the field of traffic management are based on the use of artificial intelligence. In this method, by using real-time processing of video surveillance camera images along with reinforcement learning, the optimal timing of traffic lights is determined and applied according to several parameters. In the research, deep learning methods were used in vehicle detection using the YOLOv9-C model to estimate the number and other characteristics of vehicles such as speed. Finally, by modeling vehicles in an urban environment simulator at OpenAI Gym using multi-factor reinforcement learning and the DQN Rainbow algorithm, timing is applied to traffic lights at intersections.

Additionally, the use of transfer learning along with retraining the model on images of Iranian cars has increased the accuracy of the model. The results of the proposed method show a model that is reasonably accurate in both parts of analyzing surveillance cameras and finding the optimal timing, and it has been observed that it has better accuracy than previous research.

***Keywords: Intelligent traffic management, Reinforcement learning, image processing, deep learning, traffic light management***

[1] AI researcher , +989928618623, mahdijamebozorg2000@gmail.com
[2] AI researcher, +989190981446, m.hami@eng.basu.ac.ir (Corresponding Author)
[3] Amin University instructor, sajjad.dehgaani@gmail.com